\documentclass{article}
\usepackage{ijcai18}

\usepackage{times}
\usepackage{xcolor}
\usepackage{soul}
\usepackage[utf8]{inputenc}
\usepackage[small]{caption}

\usepackage{subeqnarray}
\usepackage{amssymb}
\usepackage{color}
\usepackage{helvet}  
\usepackage{courier}  
\usepackage{url}  
\usepackage{graphicx}  

\title{Deep CNN Denoiser and Multi-layer Neighbor Component Embedding for Face Hallucination}

\author{
Junjun Jiang$^{1,2}$,
Yi Yu$^2$,
Jinhui Hu$^3$,
Suhua Tang$^4$
\and Jiayi Ma$^5$
\\
$^1$ Harbin Institute of Technology, Harbin, China\\
$^2$ National Institute of Informatics, Tokyo, Japan\\
$^3$ The Smart City Research Institute of CETC, Shenzhen, China\\
$^4$ The University of Electro-Communications, Tokyo, Japan\\
$^5$ Wuhan University, Wuhan, China\\
\{jiangjunjun, yiyu\}@nii.ac.jp,
hujinhui@cetccity.com,
shtang@uec.ac.jp,
jyma2010@gmail.com
}
\begin{document}

\maketitle
\begin{abstract}

Most of the current face hallucination methods, whether they are shallow learning-based or deep learning-based, all try to learn a relationship model between Low-Resolution (LR) and High-Resolution (HR) spaces with the help of a training set. They mainly focus on modeling image prior through either model-based optimization or discriminative inference learning. However, when the input LR face is tiny, the learned prior knowledge is no longer effective and their performance will drop sharply. To solve this problem, in this paper we propose a general face hallucination method that can integrate model-based optimization and discriminative inference. In particular, to exploit the model based prior, the Deep Convolutional Neural Networks (CNN) denoiser prior is plugged into the super-resolution optimization model with the aid of image-adaptive Laplacian regularization. Additionally, we further develop a high-frequency details compensation method by dividing the face image to facial components and performing face hallucination in a multi-layer neighbor embedding manner. Experiments demonstrate that the proposed method can achieve promising super-resolution results for tiny input LR faces.
\end{abstract}

\section{Introduction}
Face hallucination refers to the technique of reconstructing a High-Resolution (HR) face image with fine details from an observed Low-Resolution (LR) face image with the help of HR/LR training pairs~\cite{Baker2000}. It is a domain specific image super-resolution method, which focuses on the human face, and can transcend the limitations of an imaging system, thus providing very important clues about objects for criminals recognition. Due to the highly underdetermined constraints and possible noise, image super-resolution is a seriously ill-posed problem and needs the prior information to regularize the solution space. Mathematically, let $\textbf{y}$ denotes the observed LR face image, and the target HR face image $\mathbf{x}$ can be deduced by minimizing an energy function composed of a fidelity term and a regularization term balanced through a trade-off parameter $\lambda$,
\begin{equation}\label{eq:imagedegradation}
\hat \mathbf{x} = \mathop {\arg \min }\limits_{\mathbf{x}}\frac{1}{2} ||\mathbf{y}-\mathbf{Hx}||^2+\lambda \Omega (\mathbf{x}).
\end{equation}

\begin{figure}[!t]
  \centering
  \includegraphics[width=7.50cm]{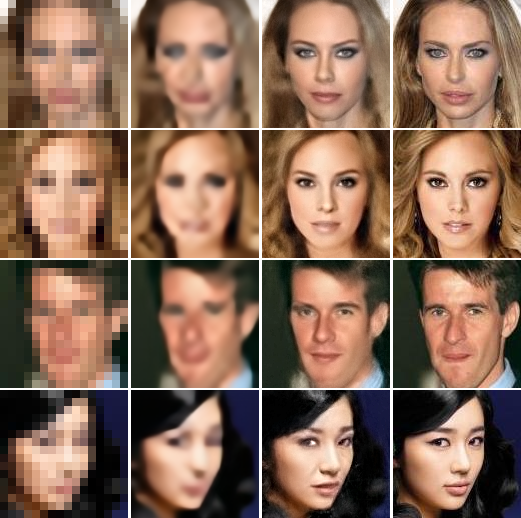}\\
   \scriptsize{{{\leftline{{\kern 21pt}16$\times$16 input{\kern 25pt} \emph{Step1}{\kern 40pt} \emph{Step2}{\kern 40pt} GT}}}}
   \vspace{-0.15cm}
  \caption{8$\times$ face hallucination results of the proposed method. \emph{Step1}: Global intermediate HR face generation via Deep CNN prior. \emph{Step2}: High-frequency face details compensation. GT: Ground truth.}\label{fig:sampleresults}
\end{figure}

\begin{figure*}
  \centering
  \includegraphics[width=16.00cm]{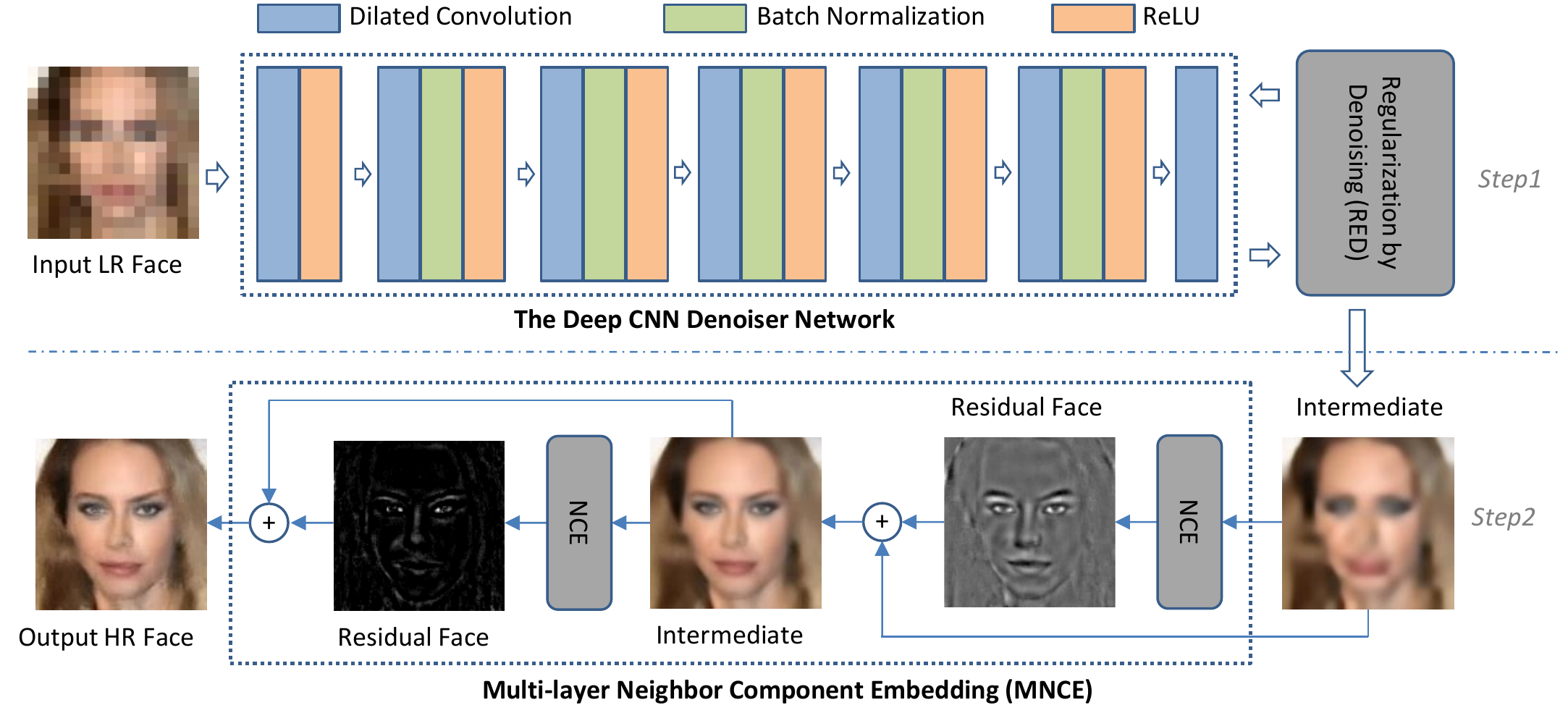}\\
  \vspace{-0.10cm}
  \caption{Main steps of the proposed face hallucination algorithm. \emph{Step1}: Deep CNN denoiser based global face reconstruction. \emph{Step2}: MNCE based residual compensation. For convenience, here we only show two layer NCE.}
  \label{fig:framework}
\end{figure*}

According to the source of the prior information of $\Omega (\mathbf{x})$, the super-resolution techniques can be divided into two categories, model-based optimization methods and discriminative inference learning methods. The former tries to solve the problem of Eq. (\ref{eq:imagedegradation}) by some time-consuming iterative optimization algorithms, while the latter aims at learning the relationship between LR and HR images through a loss function on a training set containing LR and HR sample pairs. Therefore, the model-based optimization methods (such as LRTV~\cite{shi2015lrtv} and NCSR~\cite{dong2013nonlocally}) are very general and can be used to handle various image degradation models by specifying the matrix $\mathbf{H}$. In contrast, these discriminative inference learning methods are restricted by specialized image degradation model $\mathbf{H}$. The representative discriminative learning methods include LLE \cite{Chang2004NE}, ScSR~\cite{Yang2010TIP}, ANR~\cite{timofte2013anchored}, SRCNN~\cite{dong2016image}, VDSR~\cite{Kim2016Accurate}, and some methods specifically for face images, TDN~\cite{yu2017face}, UR-DGN~\cite{yu2016ultra}, CBN~\cite{zhu2016deep}, and LCGE~\cite{song2017learning}. Due to their end-to-end training strategy, given an LR input image, they can directly predict the target HR image in an efficient and effective way.

In order to overcome the shortcomings of model-based optimization methods and discriminative inference learning methods while leveraging their respective merits, recently, some approaches have been proposed to handle the fidelity term and the regularization term separately, with the aid of variable splitting techniques, such as ADMM optimization or Regularization by Denoising (RED)~\cite{Romano2017The}. A model-based super-resolution method tries to iteratively reconstruct an HR image, so that its degraded LR image matches the input LR image, while inference learning tries to train a denoiser by machine learning, using the pairs of LR and HR images. Therefore, the complex super-resolution reconstruction problem is decomposed into a sequence of image denoising tasks, coupled with quadratic norm regularized least-squares optimization problems that are much easier to deal with. 

In many real surveillance scenarios, cameras are usually far from the interested object, and the bandwidth and storage resources of systems are limited, which generally result in very small face images, \emph{i.e.}, tiny faces. Although the above-mentioned method is general and can be used to handle various image degradation processes, the performance of this method will become very poor when the sampling factor is very large, \emph{i.e.}, the input LR face image is very small. The learned denoiser prior can not take full advantage of the structure of human face, thus the hallucinated HR faces still lack detailed features, as shown in the second column of Figure~\ref{fig:sampleresults}. In general, Deep Convolutional
Neural Networks (CNN) denoiser prior based face hallucination method generates primary face structures fairly well, but fails to return much high-frequency content. To deal with the bottlenecks of very small input images, some deep neural networks based methods have been
proposed \cite{yu2016ultra,yu2017face}.

In this paper, we develop a novel face hallucination approach via Deep CNN Denoiser and Multi-layer Neighbor Component Embedding (MNCE). Inspired by the work of \cite{ZhangKai2017Learning}, we adopt CNN to learn the denoiser prior, which is then plugged into a model-based optimization to jointly benefit the merits of model-based optimization and discriminative inference. In this step, we can predict the intermediate results, which look smooth, by this Deep CNN denoiser. In order to enhance the detailed feature, we further propose a residual compensation method through MNCE. It extends NCE to a multi-layer framework to gradually mitigate the inconsistency between the LR and HR spaces (especially when the factor is very large), thus compensating for the missing details that have not been recovered in the first step. Figure \ref{fig:framework} shows the pipeline of the proposed algorithm.

The contributions of this work are summarized as follows: (i) We proposed a novel two-step face hallucination method which combines the benefits of model-based optimization and discriminative inference Learning. The proposed framework makes it possible to learn priors from different sources (i.e., general and face images) to simultaneously regularize face hallucination. (ii) To recover the missing detailed features, neighbor component embedding with multi-layer manner is proposed, and the hallucinated result can be gradually optimized and improved. It provides a scheme to mitigate the inconsistency between LR and HR spaces due to one-to-many mappings.

\section{Related Work}
There have been several attempts to incorporate advanced denoiser priors into general inverse problems. In \cite{danielyan2012bm3d}, BM3D denoising \cite{dabov2007image} is adapted to the inverse problem of image deblurring. It was later extended by \cite{zhang2014group} to other image restoration problems. Most recently, Zhang et al. \cite{ZhangKai2017Learning} take advantage of Deep CNN discriminative learning and incorporated it to the model-based optimization methods to tackle with the inverse problems. It exhibits powerful prior modeling capacity. When the magnification is large, however, these denoiser prior based super-resolution methods cannot reconstruct the discriminant features. Therefore, residual face compensation is needed to improve the super-resolved results.

Two-step method was first proposed by Liu et al.~\cite{Liu2001}, in where the PCA based parametric model is used to generate the global face image and the MRF based local nonparametric model is adopted to compensate the lost face details in the first step. Manifold alignment based two-step methods \cite{Huang2010CCA} have been proposed to predict the target HR face image in the aligned common space. In \cite{song2017learning}, a component generation and enhancement is proposed. They firstly divided the LR test image into five facial components and obtained the basic structure by several parallel CNNs, and then fine grained facial structures are predicted by a component enhancement method.

\begin{figure}
  \centering
  \includegraphics[width=8.680cm]{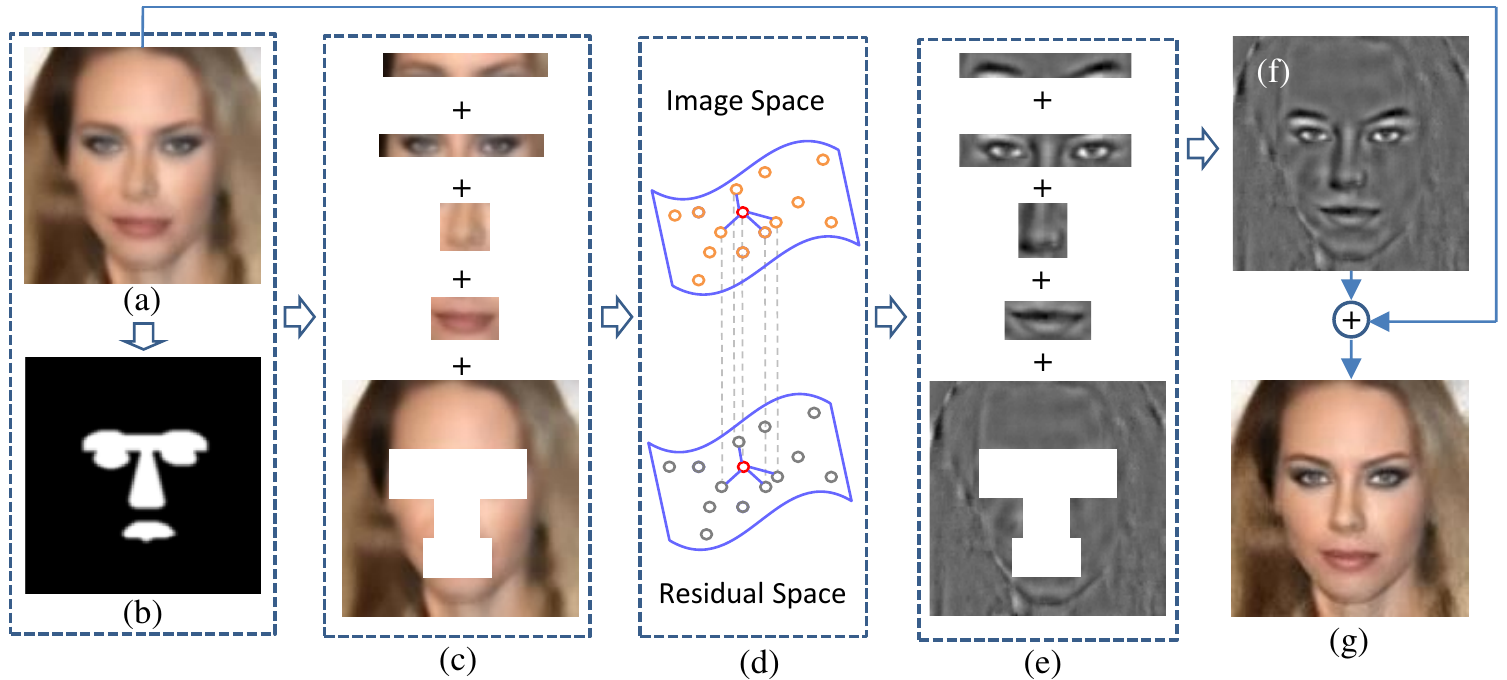}\\
  \vspace{-0.10cm}
  \caption{Illustration of neighbor component embedding based residual compensation. (a) Input image. (b) Face component masks. (c) Five facial components. (d) Neighbor embedding on the image and residual manifold spaces. (e) Constructed residual components. (f) Residual face image. (g) Hallucinated face image.}
\label{fig:NCE}
\end{figure}

\section{Proposed Algorithm}
Our precise pipeline (as shown in Figure \ref{fig:framework}) works in the following two steps. Firstly, we construct a discriminative denoiser based on the Deep CNN model. Acquiring the denoiser, the super-resolution reconstruction problem can be iteratively solved by Deep CNN denoising and RED with an image-adaptive Laplacian regularizer \cite{Milanfar2013A}. The output of this step, one intermediate HR face image, suffers from lacking detailed face features (as shown in the second column of Figure \ref{fig:sampleresults}). Secondly, we propose an MNCE based residual compensation to predict the missing detailed residual face image gradually.

\subsection{Deep CNN Denoiser Prior for Global Face Reconstruction}
\subsubsection{Regularization by Denoising for the Inverse Problem}
To solve the problem of (\ref{eq:imagedegradation}), some methods have been proposed by transforming it to an image denoising task based on some variable splitting techniques, such as ADMM optimization~\cite{boyd2011distributed,afonso2010fast} or RED based framework \cite{Romano2017The}. Since the latter adopts a theoretically better founded method than the ADMM optimization, in this paper we apply the RED to handle the restoration task (\ref{eq:imagedegradation}). In RED, the regularizer $\Omega (\mathbf{x})$ is defined by a denoiser,
\begin{equation}\label{eq:denoiser}
\hat \mathbf{x} = \mathop {\arg \min }\limits_{\mathbf{x}}\frac{1}{2} ||\mathbf{y}-\mathbf{Hx}||^2+ \frac{\lambda}{2} \mathbf{x}(\mathbf{x}-h(\mathbf{x})),
\end{equation}
where the function $h(\cdot)$ is an arbitrary denoiser. In Eq. (\ref{eq:denoiser}), the second term is an image-adaptive Laplacian regularizer \cite{Milanfar2013A}, which can lead to  either a small inner product between $\mathbf{x}$ and the residual $(\mathbf{x} - h(\mathbf{x}))$, or a small residual image. Now, the problem is how to optimize the energy function:
\begin{equation}\label{eq:energy}
\mathbf{E}(\mathbf{x}) = \frac{1}{2} ||\mathbf{y}-\mathbf{Hx}||^2+ \frac{\lambda}{2} \mathbf{x}(\mathbf{x}-h(\mathbf{x})).
\end{equation}
Following \cite{Romano2017The}, which states that the gradient of $\Omega (\mathbf{x})$ can be induced under the mild assumptions, \emph{i.e.}, $\nabla_{\mathbf{x}} \Omega (\mathbf{x}) =\mathbf{x}-h(\mathbf{x})$. Thus, we can obtain the gradient of $\mathbf{E}(\mathbf{x})$ by
\begin{equation}\label{eq:gradient}
\nabla_{\mathbf{x}} \mathbf{E} (\mathbf{x}) = \mathbf{H}^T(\mathbf{H}\mathbf{x}-\mathbf{y})+\lambda (\mathbf{x}-h(\mathbf{x})).
\end{equation}
Therefore, we can easily get the update rule by setting $\nabla_{\mathbf{x}} \mathbf{E} (\mathbf{x}) =0$,
\begin{equation}\label{eq:gradient}
\begin{array}{l}
0= \mathbf{H}^T(\mathbf{H}\hat \mathbf{x}_{k+1}-\mathbf{y})+\lambda (\hat \mathbf{x}_{k+1}-h(\hat \mathbf{x}_{k}))\\
\\
\Rightarrow \hat \mathbf{x}_{k+1}= (\mathbf{H}^T\mathbf{H}+\lambda \mathbf{I})^{-1}(\mathbf{H}^T\mathbf{y}+\lambda h(\hat \mathbf{x}_{k})).
 \end{array}
\end{equation}

Through a sequence of image denoising problems and $L_2$ norm regularized least-squares optimization problems, we can take full advantage of model-based optimization methods and discriminative inference learning methods: various degradation process can be handled and advanced denoiser prior can be easily incorporated.

\subsubsection{Learning the Deep CNN Denoiser Prior}
Inspired by \cite{ZhangKai2017Learning}, we also introduce the Deep CNN denoiser to model the discriminative image prior for its efficiency sue to parallel computation ability of GPU and powerful prior modeling capacity with deep neural networks . The above part of Figure \ref{fig:framework} shows the architecture of the Deep CNN denoiser network, which consists of seven hidden layers, ``Dilated Convolution + ReLU'' block in the first layer, five ``Dilated Convolution + Batch Normalization + ReLU'' blocks in the middle layers, and ``Dilated Convolution'' block in the last layer.

Once the network is trained, we can predict the result by iterative Deep CNN based denoising and solving the $L_2$ norm regularized least-squares optimization problem. From previous discussion, we learn that this method will become very poor and fail to return much high-frequency content when the sampling factor is very large, due to ignoring the structure of human face, which is a highly structured object. In the following, we will introduce an improvement method to enhance the high-frequency content.

\begin{figure}[!t]
  \centering
     \includegraphics[width=8.50cm]{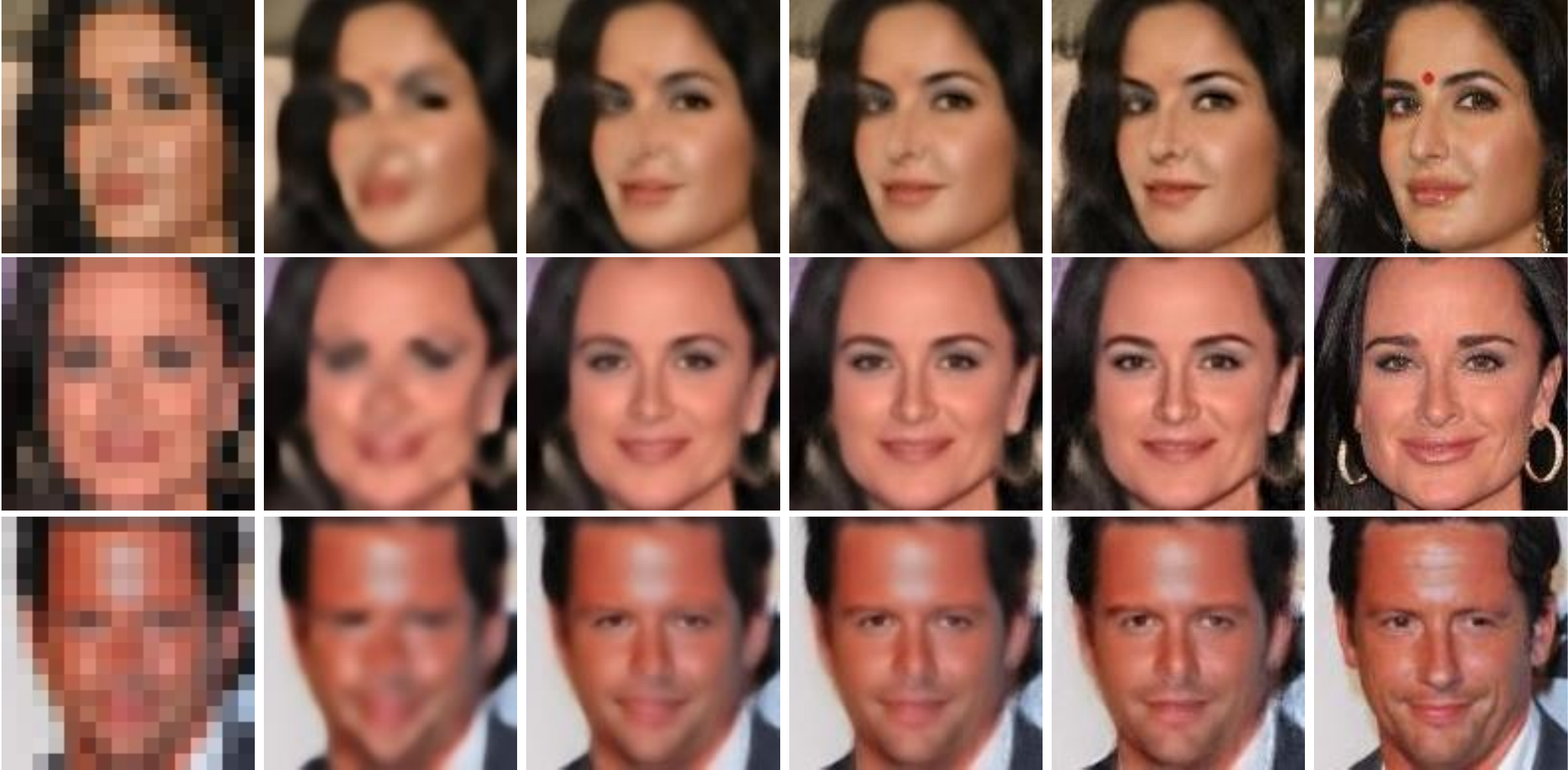}\\
     \scriptsize{ {\kern 0pt}(a){\kern 30pt} (b) {\kern 30pt} (c) {\kern 30pt} (d) {\kern 30pt} (e)  {\kern 30pt} (f)}
    \vspace{-0.10cm}
    \caption{Face hallucination results of different steps of the proposed method. (a) Input. (b) \emph{Step1}. (c) \emph{Step2-Layer1}. (d) \emph{Step2-Layer2}. (e) \emph{Step2-Layer3}. (f) GT.}
\label{fig:sresultssteps}
\end{figure}

\begin{figure}[!t]
  \centering
     \includegraphics[width=8.50cm]{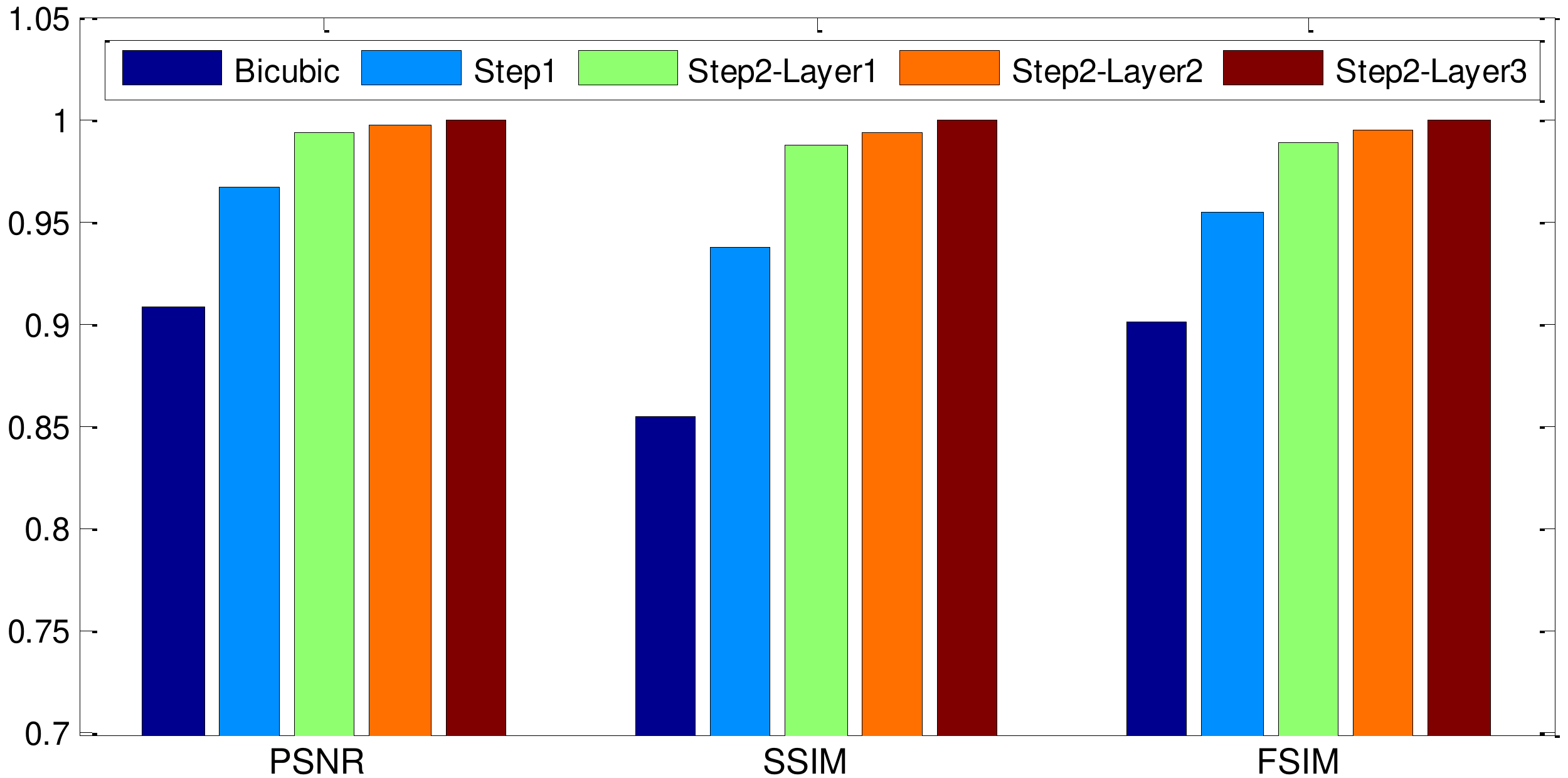}\\
     \vspace{-0.10cm}
    \caption{PSNR (dB), SSIM, and FSIM results of different steps of the proposed method. Note that we scale these three indices to [0, 1] by dividing their maximums, respectively.}

\label{fig:steps_improve}
\end{figure}

\subsection{Multi-layer Neighbor Component Embedding (MNCE) based Residual Compensation}
We take the assumption that similar LR contents will share similar potential HR contents. Let $f(\mathbf{y})$ denotes the prediction function, $\mathbf{x}-f(\mathbf{y})$ is the high-frequency residual face image. Therefore, we can construct the HR face $\mathbf{x}'$ with high-frequency residual information through the locality regularized neighbor embedding algorithm,
\begin{equation}\label{eq:weight}
\begin{array}{l}
\mathbf{x'} = f(\mathbf{y}) + \sum\limits_{k = 1}^K {{w_k^*}({\mathbf{x}_k} - f({\mathbf{y}_k}))} {\kern 10pt} where \\
 \mathbf{w^*} = \mathop {\arg \min }\limits_{\mathbf{w}}\left\| {f(\mathbf{y}) - \sum\limits_{k = 1}^K {{w_k}f({\mathbf{y}_k})} } \right\|_2^2 + \lambda \left\| {\mathbf{d} \odot \mathbf{w}} \right\|_2^2, \\
 \end{array}
\end{equation}
where $\odot$ denotes point-wise vector product, $f(\mathbf{y}_k)$ refers to $K$-nearest-neighbor (in the training set) to $f(\mathbf{y})$, $\mathbf{w}=[w_1,w_2,...,w_K]$ is the embedding weight of $f(\mathbf{y})$ from the global face space to the residual face space, and $\mathbf{d}$ is a $K$-dimensional locality adaptor that gives different freedom for each training sample, $f(\mathbf{y}_1), f(\mathbf{y}_2), ..., f(\mathbf{y}_K)$, proportional to its similarity to the input $f(\mathbf{y})$. Specifically,
\begin{equation}\label{eq:dist}
d_k=\left\|f(\mathbf{y}_k)-f(\mathbf{y})\right\|_2.
\end{equation}
In Eq. (\ref{eq:weight}), the first term represents the reconstruction error with $K$-NN, the second term represents the local geometry constraint of manifold. Here, the regularization parameter $\lambda$ represents the trade-off between the closeness to the data and the locality regularization term. Different from traditional LLE based reconstruction method \cite{Roweis2000}, which treats each $K$-NN equally, our method can give different weights to different $K$-NN, \emph{i.e.}, the dissimilar samples will be penalized heavily and obtain very small reconstruction weights, while the similar samples will be given more freedom and obtain large reconstruction weights. Thus, our method can capture salient properties as well as yield minimized reconstruction error.

\begin{figure}[!t]
  \centering
     \includegraphics[width=8.50cm]{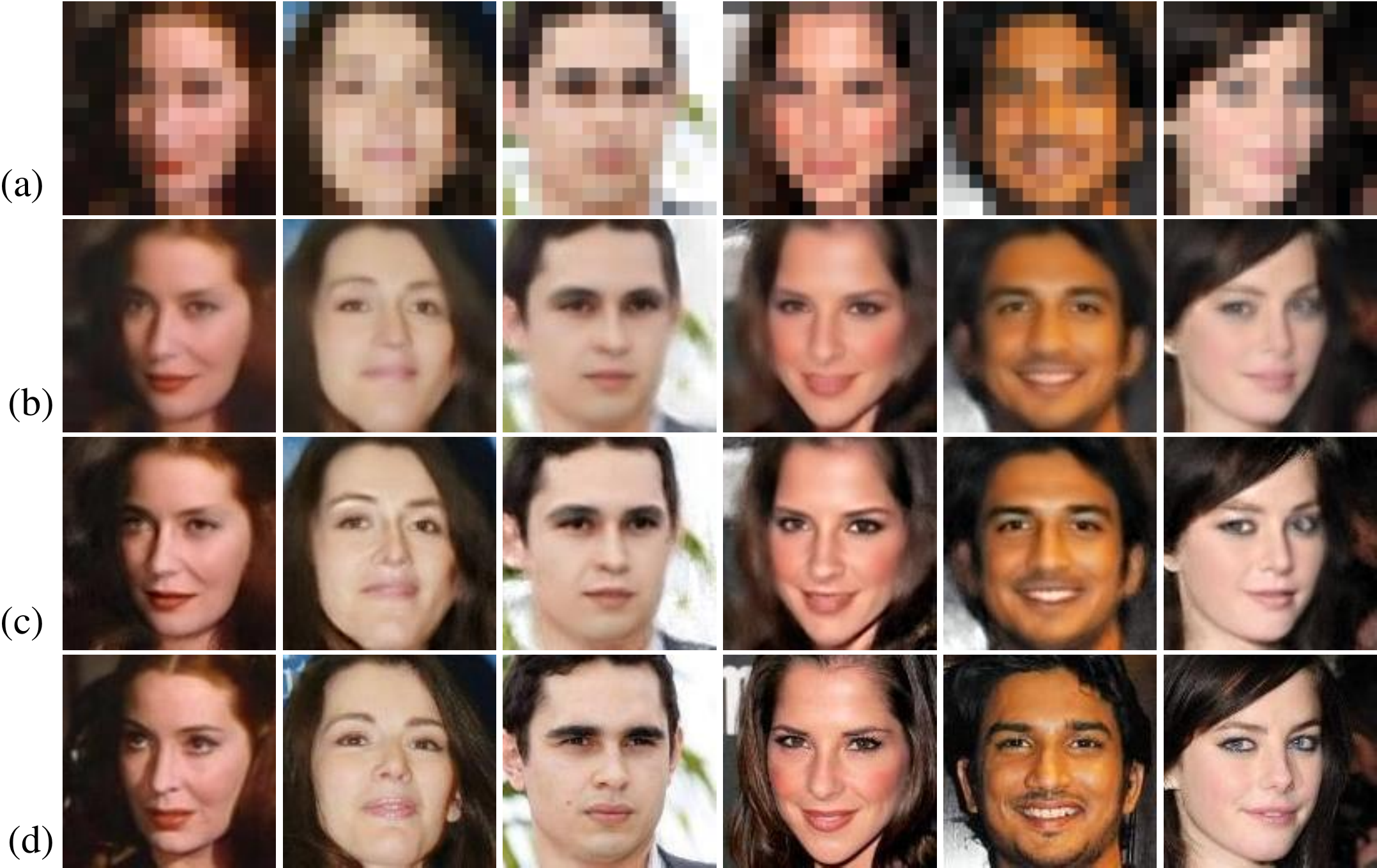}\\
     \vspace{-0.17cm}

    \caption{Visual comparisons when using different global face reconstruction methods. (a) Input. (b) Bicubic + MNCE. (c) Deep Denoiser + MNCE. (d) GT.}

\label{fig:BI_MNCE}
\end{figure}

\begin{figure*}[!t]
  \centering
  \includegraphics[width=17.70cm]{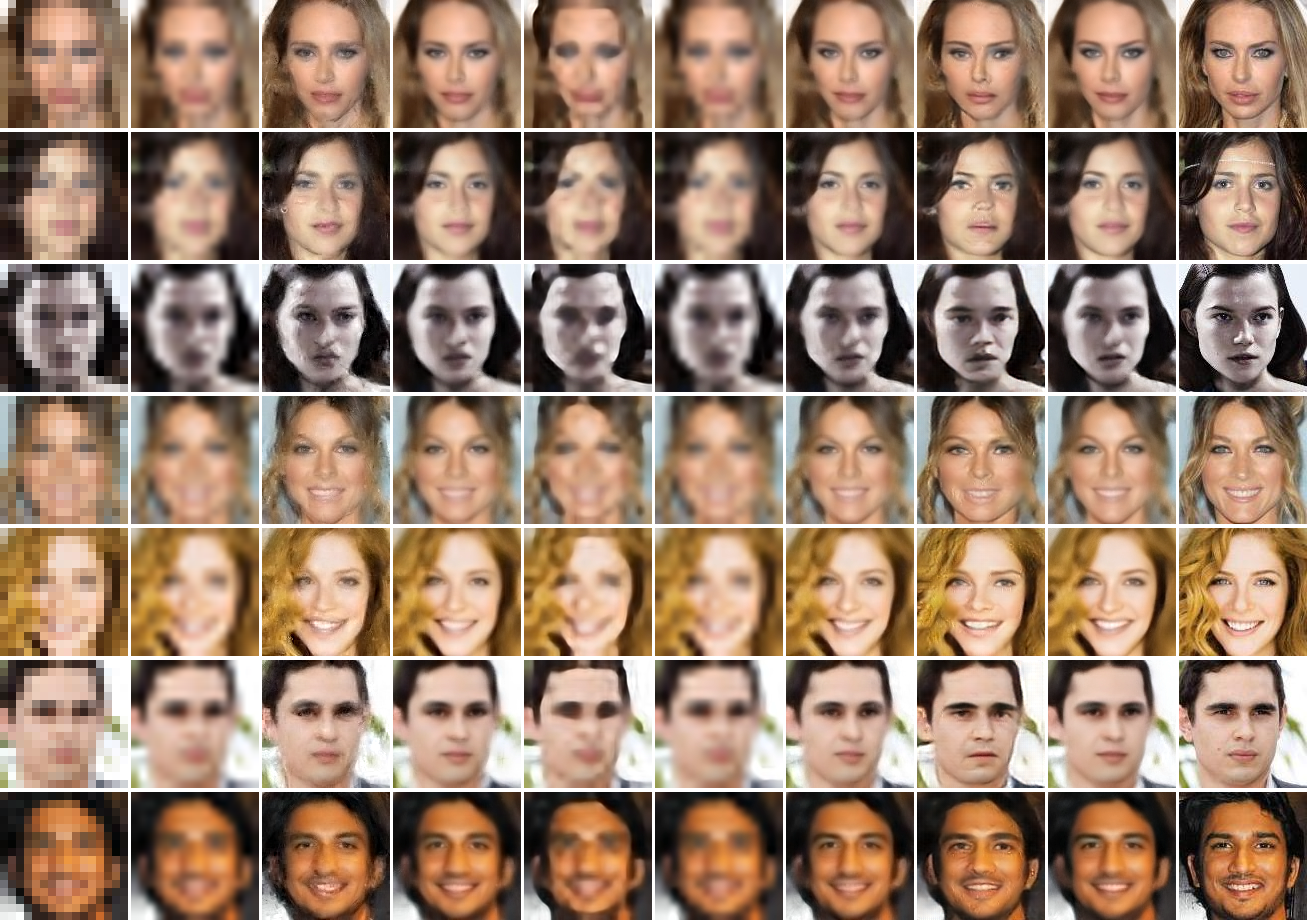}\\
 \scriptsize{\leftline{{\kern 8pt}16$\times$16 Input{\kern 19pt} Bicubic{\kern 32pt} LLE{\kern 35pt} LcR {\kern 30pt} SRCNN{\kern 27pt} VDSR{\kern 30pt} LCGE{\kern 27pt} UR-DGN{\kern 28pt} Our {\kern 38pt} GT}}
  \vspace{-0.18cm}
  \caption{8$\times$  face hallucination comparisons with state-of-the-arts on near frontal input faces. Please zoom in to see the differences.}
\label{fig:frontal}
\end{figure*}

\begin{figure*}[!t]
  \centering
  \includegraphics[width=17.70cm]{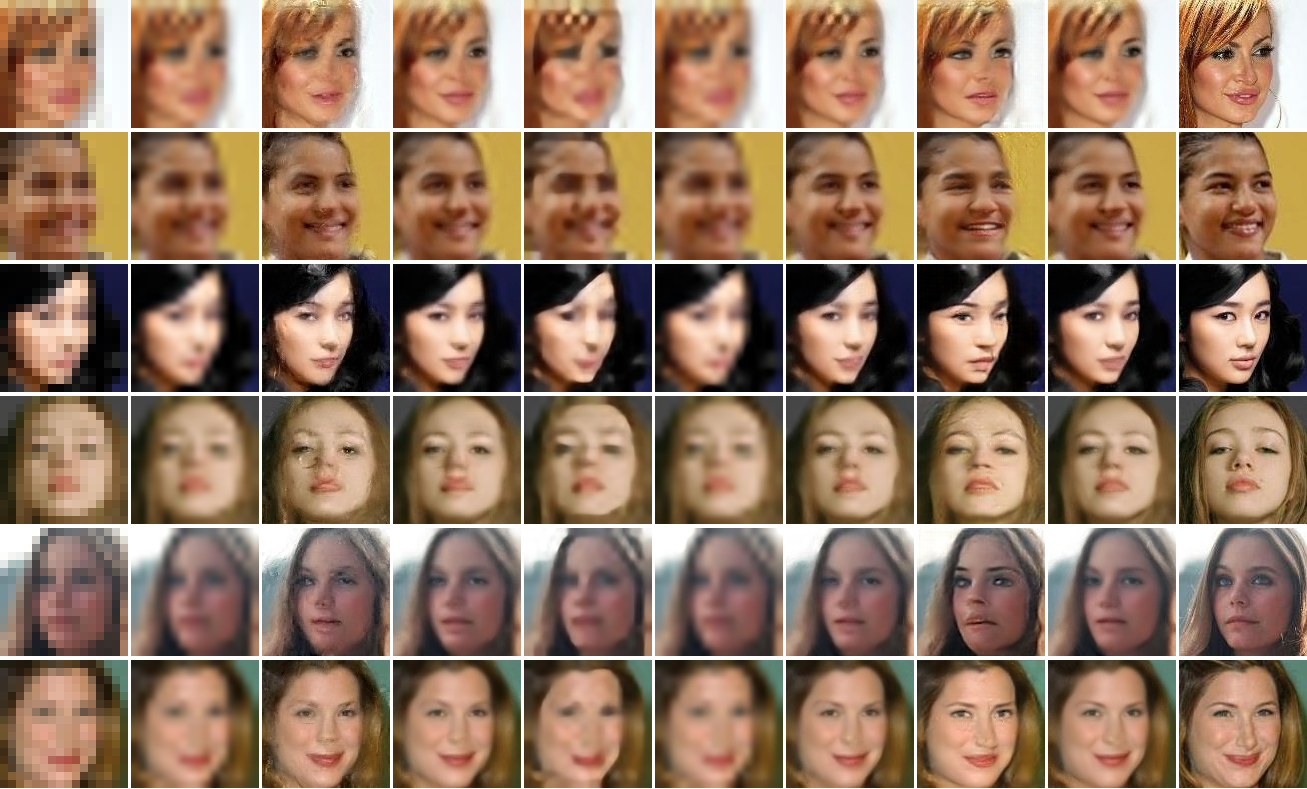}\\
 \scriptsize{\leftline{{\kern 8pt}16$\times$16 Input{\kern 19pt} Bicubic{\kern 32pt} LLE{\kern 35pt} LcR {\kern 30pt} SRCNN{\kern 27pt} VDSR{\kern 30pt} LCGE{\kern 27pt} UR-DGN{\kern 28pt} Our {\kern 38pt} GT}}
 \vspace{-0.18cm}
  \caption{8$\times$  face hallucination comparisons with state-of-the-arts on non-frontal input faces. Please zoom in to see the differences.}
\label{fig:non-frontal}
\end{figure*}

\subsubsection{Neighbor Component Embedding}
The above method is limited to reconstruct the entire high-frequency faces, but it is hard for us to find the entire faces that are very similar to the input one. Similar to \cite{song2017learning,Yang2013CVPR}, we also divide a face image into five components, \emph{e.g.}, eyes, eyebrows, noses, mouths, and the remaining region, as shown in Figure~\ref{fig:NCE}(c). By dividing a face image into different components, we can embed each component from the image space to the residual component face space separately,
\begin{equation}\label{eq:weight2}
\begin{array}{l}
\mathbf{x'}_j = f_j(\mathbf{y}) + \sum\limits_{k = 1}^K {{w_{jk}^*}({\mathbf{x}_{jk}} - f_j({\mathbf{y}_{k}}))}, {\kern 3pt}  \rm{where} \\
 \mathbf{w^*}_j = \mathop {\arg \min }\limits_{\mathbf{w}_j}\left\| {f_j(\mathbf{y}) - \sum\limits_{k = 1}^K {{w_{jk}}f_j({\mathbf{y}_{k}})} } \right\|_2^2 + \lambda \left\| {\mathbf{d}_j \odot \mathbf{w}_j} \right\|_2^2, \\
 \end{array}
\end{equation}
where $\mathbf{x}_{jk}$, $f_j({\mathbf{y}_{k}})$, and $f_j({\mathbf{y}})$ are the $j$-th component of $\mathbf{x}_{k}$, $f({\mathbf{y}_{k}})$, and $f({\mathbf{y}})$, respectively, and $\mathbf{w}_j$ is the corresponding embedding vector of $f_j({\mathbf{y}})$.  Illustration of neighbor component embedding is given in Figure \ref{fig:NCE}. For each facial component, we transform it from the LR image space to the residual image space by neighbor embedding. In this way, the high-frequency residual face (Figure \ref{fig:NCE}(f)) can capture tiny details.

\subsubsection{Multi-layer Embedding Enhancement}
From previous works, we learn that the similar local manifold structure assumption of LR and HR spaces is not always holden in practice. As reported in \cite{Jiang2014TIP}, the neighborhood preservation rates decrease with the increase of downsampling factor or noise level. In order to reduce the gap between the LR and HR manifold spaces, we introduce a multi-layer embedding enhancement based on the observation that the reconstructed HR manifold of the LR training samples is much more consistent than that of the original LR manifold. With the reconstructed HR training samples and the corresponding HR training samples, we can perform super-resolution reconstruction in much more consistent coupled LR and HR spaces. Specially, in the training phase, we can leverage the ``leave-one-out'' strategy to obtain the global face based on Deep CNN denoiser, and then predict the residual face through neighbor component embedding for all the LR training face images. When all the LR training face image are updated (super-resoved), we generate a new ``LR'' training set and take it as the input of the next neighbor embedding layer. In the testing phase, the input LR face can be gradually super-resolved to a satisfactory result.

\section{Experiments}
The performance of the proposed algorithm has been evaluated on the large-scale Celebrity Face Attributes (CelebA) dataset~\cite{liu2015deep}, and we compared our method with the state-of-the-arts qualitatively and quantitatively on the dataset. We adopt the widely used Peak Signal-to-Noise Ratio (PSNR), structural similarity(SSIM) \cite{Wang2004SSIM} as well as feature similarity (FSIM) \cite{Zhang2011FSIM} as our evaluation measurements.

\subsection{Dataset}
We use the Celebrity Face Attributes (CelebA) dataset \cite{liu2015faceattributes} as it consists of subjects with large diversities, large quantities, and rich annotations, including 10,177 identities and 202,599 face images. We select ten percent of the data, which includes 20K training images and 260 testing images. And then, these images are aligned and cropped to 128$\times$128 pixels as HR images. The LR images are obtained by Bicubic 8$\times$ downsampling (default setting of Matlab function imresize), and thus the input LR faces are 16$\times$16 pixels.

\subsection{Effectiveness of the Proposed Two-step Methods}
To demonstrate the effectiveness of the proposed two-step methods, we give the intermediate results of different steps. As shown in Figure \ref{fig:sresultssteps}, by performing the Deep CNN denoiser based global face reconstruction (\emph{Step1}), it can well maintain the primary facial contours. Through layer-wise component embedding (\emph{Step2}), we can expect to gradually enhance the characteristic details of the reconstructed results (please refer to the third to the fifth columns). As a learned general prior, the Deep CNN denoiser prior cannot be used to model the facial details. However, it can be used to mitigate the manifold inconsistence between the LR and HR image spaces, and this will benefit the following neighbor component embedding learning. At the second step, it is much easier to predict the relationship between the LR and HR spaces when the gap of manifold structure between them is small. Figure \ref{fig:steps_improve} quantitatively shows the effectiveness of multi-layer embedding. It demonstrates that by iteratively embedding, we can expect to gradually approach the ground truth.

To demonstrate the effectiveness of the Deep CNN denoiser based global face reconstruction model, we further show the hallucination results of replacing Deep CNN denoiser based global face reconstruction with Bicubic interpolation while keeping the second step (\emph{i.e.}, MNCE) as the same. As shown in Figure \ref{fig:BI_MNCE}, Deep CNN denoiser can produce clearer and shaper facial contours. In addition, we also noticed that Bicubic with MNEC can also infer reasonable results, which verifies the ability of MNCE when learning the relationship between the LR faces and residual images.

\subsection{Qualitative and Quantitative Comparisons}
We compare our method with several representative methods, which include LLE \cite{Chang2004NE} and LcR \cite{Jiang2014LcRTMM}, two representative deep learning based methods, SRCNN~\cite{dong2016image}, VDSR~\cite{Kim2016Accurate}, and two most recently proposed face specific image super-resolution methods, \emph{i.e.}, LCGE~\cite{song2017learning} and UR-DGN \cite{yu2016ultra}. Bicubic interpolation in also introduced as a baseline.

\begin{figure}[!t]
  \centering
  \includegraphics[width=8.50cm]{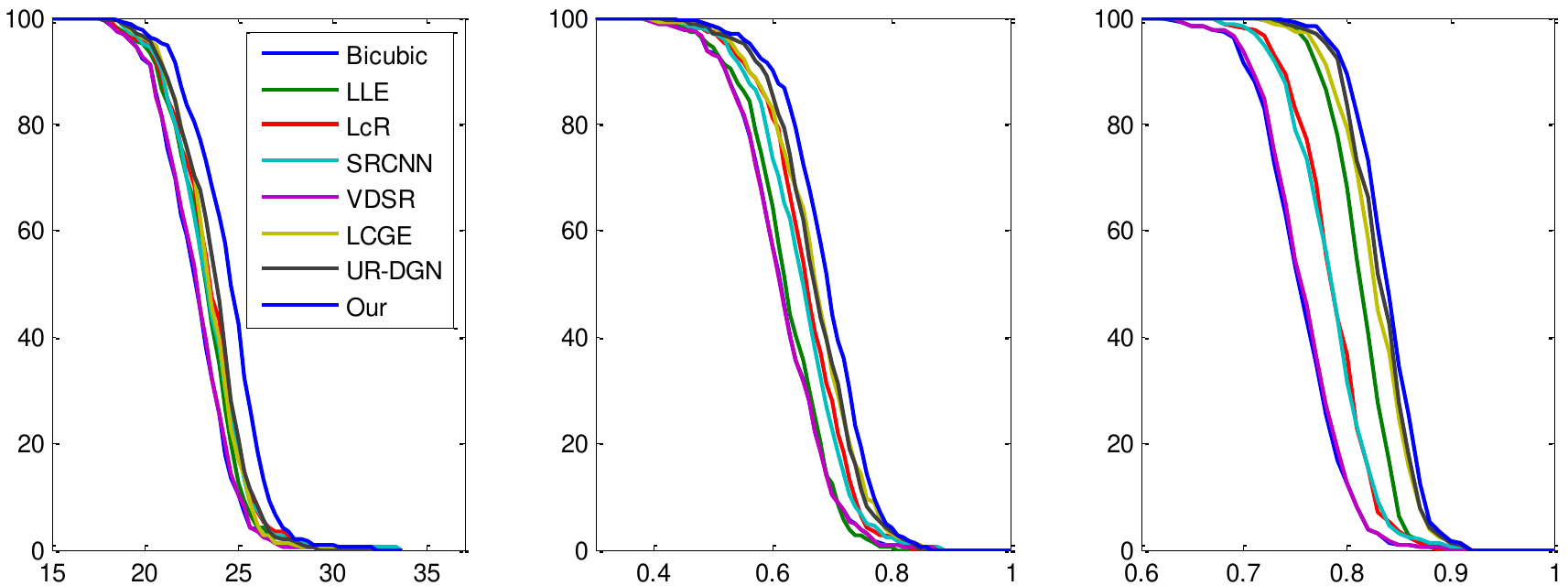}\\
  \scriptsize{(a) PSNR{\kern 55pt}  (b) SSIM{\kern 55pt} (c) FSIM}
   \vspace{-0.120cm}
  \caption{Image quality statistics using (a) PSNR (dB), (b) SSIM, and (c) FSIM. The horizontal axis labels the scores using PSNR, SSIM, or FSIM, while the vertical axis marks the percentage of hallucinated HR face images whose scores are larger than the score marked on the horizontal axis.}

  \label{fig:three}
\end{figure}

\setlength{\tabcolsep}{3.3pt}
\begin{table}
\centering
\scriptsize
\begin{tabular}{|c|c|c|c|c|c|c|c|c|}
\hline
Index   &   Bicubic &   LLE   &   LcR   &   SRCNN   &   VDSR   &   LCGE   &   UR-DGN   &   Our\\
\hline
\hline
\emph{PSNR}   &   22.61 	&   23.08 	&   23.11 	&   23.27 	&   22.65 	&   23.35 	&   23.55  &   24.34\\
\emph{SSIM}   &   0.6134 	&   0.6208 	&   0.6542 	&   0.6463 	&   0.6128 	&   0.6673 	&   0.6696  &   0.6883\\
\emph{FSIM}   &   0.7541 	&   0.8118 	&   0.7843 	&   0.7828 	&   0.7558 	&   0.8257 	&   0.8309   &   0.8375\\
\hline
\end{tabular}
\caption{Average scores in terms of PSNR (dB), SSIM, and FSIM of different face hallucination approaches.}
\label{tab:scores}
\end{table}

As shown in Figure \ref{fig:frontal}, we also compare the visual results of different comparison methods. 
It shows that the basic Bicubic interpolation method cannot produce additional details, whereas LLE may introduce some high frequency that doesn’t exist. LcR, which focuses on the well aligned frontal face reconstruction, will inevitably smooth the final result due to the misalignments between training samples. As for the deep learning based technologies, such as SRCNN and VDSR, they can well maintain the face contours due to their global optimization scheme. However, they fail to capture high frequency details (please refer to the eyes, noses, and mouth). This is mainly because when the magnification factor is large, it is very difficult for them to learn the relationship between the LR and HR images with an end-to-end manner. As a gradual super-resolution approach, LCGE method and the proposed method can infer the original low-frequency global face structure as well as the high-frequency local face details simultaneously. When we look further at the results of LCGE and the proposed method, we learn that our method can produce clearer HR faces (please refer to the eyes, mouths, and facial contours). When compared with UR-DGN, which can be seen as the current most competitive face hallucination method for tiny input, our results are still very competitive and much more reasonable. UR-DGN achieves relatively sharper face contours, but the hallucinated faces are dirty.

In addition to the results on near frontal faces (Figure \ref{fig:frontal}), in Figure \ref{fig:non-frontal} we also show some visual hallucination results with non-frontal faces, to further demonstrate the robustness of the proposed method. The advantages of the proposed method are still obvious, especially for the regions of eyes and mouth. For examples, the resultant faces of LcR, SRCNN, and VDSR lack detailed information, LLE introduces some unexpected high-frequency details, and UR-DGN may produce sharp but dirty faces. Although the same for component embedding based two-step method, the proposed method is much more robust to pose variety than the approach of LCGE.

Figure \ref{fig:three} shows the statistical curves of PSNR (dB), SSIM, and FSIM scores of different face hallucination approaches, and Table \ref{tab:scores} tabulates their average scores. It shows a considerable quantitative advantage of our method compared to traditional shallow learning based methods and some recently proposed deep learning based methods. By comparing UR-DGN and our method, we learn that the proposed method can generate more reliable results, while UR-DGN can well maintain structure information but introduce dirty pixels.

\subsection{Face Hallucination with Surveillance Faces}
\begin{figure}[!t]
  \includegraphics[width=8.60cm]{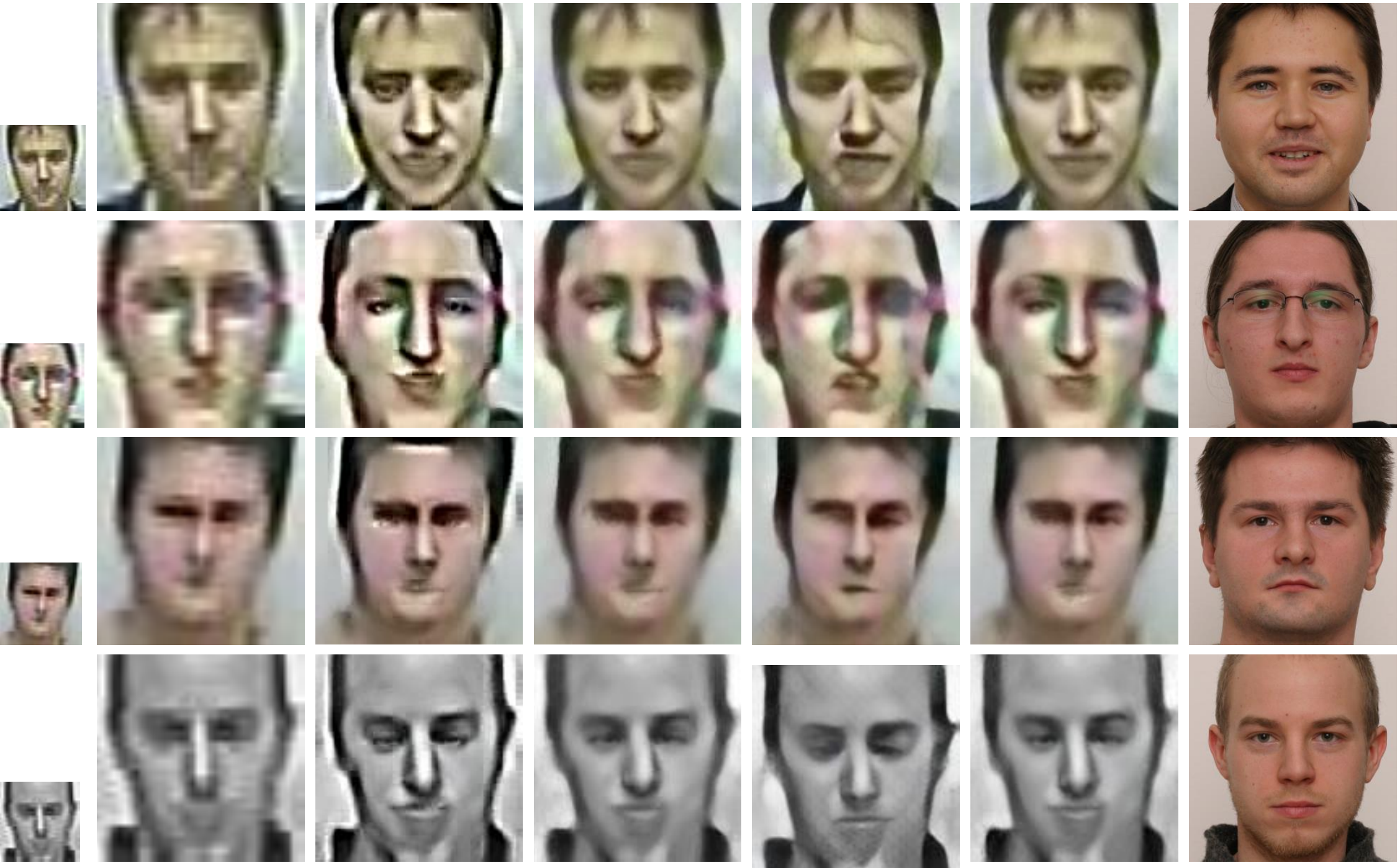}\\
   \scriptsize{ {\kern 8pt}LR{\kern 14pt} Bicubic {\kern 17pt} LcR {\kern 20pt} LCGE {\kern 15pt} UR-DGN {\kern 14pt} Our  {\kern 15pt} Reference}
   \vspace{-0.120cm}
  \caption{Real-world face hallucination results of different approaches with low-quality surveillance face images. }
  \label{fig:realresults2}
\end{figure}

While existing methods can perform well on standard test databases, they often perform poorly when they encounter low-quality and LR face images obtained in real-world scenarios.
Figure \ref{fig:realresults2} shows some face hallucination results on the SCface dataset \cite{Grgic2011SCface}
in which images mimic the real world conditions. The first column is the input LR face image, while the last is the reference HR face image of the corresponding individual that can be seen as the ground truth. The middle four columns are the results of LcR, LCGE, UR-GDN, and the proposed method. We observe that these results are obviously worse than those under the CelebA dataset, which shows the shortcomings of learning based methods that require statistical consistency between the training and testing samples. For example, for the eye regions of the hallucinated results, there are more artifacts than the results in the CelebA dataset. This is mainly due to the self-occlusion problem caused by the pose (\emph{e.g.}, looking down) of surveillance cameras, and it is hard to find such samples in a standard face dataset like CelebA.

\section{Conclusions and Future Work}
In this paper, we presented a novel two-step face hallucination framework for tiny face images. It jointly took into consideration the model-based optimization and discriminative inference, and presented a Deep CNN denoiser prior based global face reconstruction method. And then, the global intermediate HR face was gradually embedded into the HR manifold space with a multi-layer neighbor component embedding manner. Empirical studies on the large scale face dataset and real-world images demonstrated the effectiveness and robustness of the proposed face hallucination framework.

The input faces are aligned manually or by other algorithms. In future work, we need to consider the face alignment and parsing to hallucinate an LR face image with unknown and arbitrary poses \cite{zhu2016deep,chen2017fsrnet,Yu2018CVPR}.

\section*{Acknowledgments}
The research was supported by the National Natural Science Foundation of China under Grants 61501413 and 61503288, and was also partially supported by JSPS KAKENHI Grant Number 16K16058.

\footnotesize
\bibliographystyle{named}
\bibliography{ijcai18}

\end{document}